\def\year{2018}\relax
\documentclass[letterpaper]{article} 
\usepackage{alec18}  
\usepackage{times}  
\usepackage{helvet}  
\usepackage{courier}  
\usepackage{url}  
\usepackage{graphicx}  
\usepackage{subcaption}
\usepackage{tabularx}
\begin{document}
%
\title{Projecting Trouble: Light Based Adversarial Attacks on Deep Learning Classifiers}
\author{
	Nicole Nichols \textsuperscript{1,2}\\
	\texttt{nicole.nichols@pnnl.gov} \\
	\textsuperscript{1}Pacific Northwest National Laboratory, Seattle, Washington\\
	\textsuperscript{2}Western Washington University, Bellingham, Washington\\
	\And
	Robert Jasper\textsuperscript{1}\\
	\texttt{robert.jasper@pnnl.gov}\\
}




\maketitle
\begin{abstract}
This work demonstrates a physical attack on a deep learning image classification system using projected light onto a physical scene. 
Prior work is dominated by techniques for creating adversarial examples which directly manipulate the digital input of the classifier. 
Such an attack is limited to scenarios where the adversary can directly update the inputs to the classifier.
This could happen by intercepting and modifying the inputs to an online API such as Clarifai or Cloud Vision. 
Such limitations have led to a vein of research around physical attacks where objects are constructed to be inherently adversarial or adversarial modifications are added to cause misclassification. 
Our work differs from other physical attacks in that we can cause misclassification dynamically without altering physical objects in a permanent way. 

We construct an experimental setup which includes a light projection source, an object for classification, and a camera to capture the scene. 
Experiments are conducted against 2D and 3D objects from CIFAR-10. Initial tests show projected light patterns selected via differential evolution could degrade classification from 98\% to 22\% and 89\% to 43\% probability for 2D and 3D targets respectively. Subsequent experiments explore sensitivity to physical setup and compare two additional baseline conditions for all 10 CIFAR classes. Some physical targets are more susceptible to perturbation. Simple attacks show near equivalent success, and 6 of the 10 classes were disrupted by light. 

\end{abstract}

\section{Introduction}
Machine learning models are vulnerable to adversarial attacks by making small but targeted modifications to inputs that cause misclassification.
The research around adversarial attacks on deep learning systems has grown significantly since \cite{szegedy2013intriguing} demonstrated intriguing properties.
The scope and limitations of such attacks is an active area of research in the academic community.  
Most of the research has focused on the purely digital manipulation.
Recently, researchers have developed techniques that alter or manipulate physical objects to fool classifiers, which could pose a much greater real world threat.

\section{Related Research}\label{RR}
Researchers have proposed many theories about the cause of model vulnerabilities. Evidence suggests that adversarial samples lie close to the decision boundary in the low dimensional manifold representing high dimensional data. 
Adversarial manipulation in the high dimension is often imperceptible to humans and can shift the low dimensional representation to cross the decision boundary \cite{feinman2017detecting}. 
Many approaches are available to perform this manipulation if the attacker has access to the defender's classifier. 
Furthermore, adversarial examples have empirically been shown to transfer between different classifier types \cite{Papernot2016,szegedy2013intriguing}. 
This enhances the attacker's potential capability when there is no access to the defender's classifier. 

It is difficult for defenses to keep pace with attacks, and the advantage lies with the adversary. 
This was highlighted when seven of the eight white box defenses announced at the prestigious ICLR2018 were defeated within a week of publication \cite{athalye2018obfuscated}.

Researchers have successfully demonstrated physical world attacks against deep learning classifiers. 
Some of the first physical attacks were demonstrated by printing an adversarial example, photographing the printed image, and verifying the adversarial attack remained \cite{Kurakin2016}. 
\cite{sharif2016accessorize} demonstrated printed eyeglasses frames that thwart facial recognition systems and fully avoid face detection by the Viola-Jones object detection algorithm. 
It has also been noted that near infra-red light can also be used to evade face detection \cite{yamada2013privacy}. 
Our work is different because we leverage dynamic generation methods use real world feedback when learning the patterns of light to project. 
 
Putting aside adversarial attacks, most image classifiers are not inherently invariant to object scale, translation, or rotation. 
Notable exceptions are \cite{cohen2014transformation}, which attempts to learn object recognition by construction of parts, and \cite{qi2017pointnet} which use 3D point cloud representation for object classification. 
To some degree, this invariance can be learned from training data if it has intentionally been designed to address this gap. For example the early work by \cite{lecun2004learning} was evaluated with the NORB dataset which was systematically collected to assess pose, lighting, and rotation of 3D objects.

Simulating scale, translation, and rotation of 2D images is conducive to experiment  automation, and many recent advances in rotational invariance such as Spatial Transformer Networks \cite{jaderberg2015spatial}, use this framework for evaluation of robustness to these properties. 
However, further research is needed to validate the ability of this simulated rotational invariance to transfer to real world rotation of 3D figures. 
We emphasize the need for invariant models because it is impossible to disambiguate the success of an attack when it is can only be validated with a weak model.  

Maintaining adversarial attack under a range of pose or lighting conditions may prove to be the most difficult aspect of this task. 
Some preliminary research suggests this is possible and demonstrated two toy examples in the physical world \cite{athalye2017synthesizing}. 
They introduce an Expectation over Transformation (EoT) method for differentiating texture patterns through a 3D renderer to produce an adversarial object. 
An additional demonstration of physical attack is to introduce an adversarial patch to the physical scene, which is invariant to location, rotation, scale, and cause specific misclassification \cite{brown2017adversarial}.


\section{Experimental Setup and Results}
We constructed a test environment to perform light based adversarial attacks and collect data in an office environment with minimal lighting control. 
Our attacks were conducted against 2D and 3D target objects placed in the scene. 
We used a projector to project light onto the target and a common web camera to capture the scene. 
For the 2D and initial 3D experiments, the projector was a Casio XJ-A257 and the camera was a Logitech C930e. 
During the second phase of 3D experiments, we used an Epson VS250 projector, Logitech C615 HD camera and an Altura HD-ND8, neutral density filter to control the light intensity of the projector. 

\subsection{2D Presentation}
\begin{figure*}[]
	\centering
	\begin{subfigure}[]{0.45\textwidth}
		\centering
		\includegraphics[width=.45\textwidth]{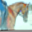}
		\caption{The 2D scene without adversarial attack.}\label{fig:horse}
	\end{subfigure}
	~
	\begin{subfigure}[]{0.45\textwidth}
		\centering
		\includegraphics[width=.45\textwidth]{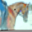}
		\caption{The 2D scene with adversarial attack.}\label{fig:horse-attack}
	\end{subfigure}
	\caption{Images demonstrating light based attack on 2D physical presentation}\label{2D attack}
\end{figure*}

For the 2D scene, we chose a random image (\texttt{horse}) from the CIFAR-10 dataset to be attacked. 
The image was printed and secured to the wall in front of the camera and projector. 
Following a similar methodology of earlier work \cite{su2017one} on single pixel attacks we use differential evolution (DE) to optimize a light based attack to cause misclassification. 
Differential evolution is a heuristic global optimization strategy similar to genetic algorithms where the algorithm maintains a population of candidate solutions, selecting a small number (potentially one) for further rounds of modification and refinement.
We projected a digital black 32x32 square containing a single pixel at a variable location and RGB values. 
Because projectors can't project black (the absence of light) the projector adjusted the black pixels to present the illusion of a black background. 
This adjustment is impacted somewhat by RGB value of the single pixel being projected.
Each iteration of the differential evolution was projected, captured, and input to a standard ResNet38 for classification of the image captured by the camera.
Though only one pixel was modified in the digital attack pattern, because of the distance between the projector and object, a larger area in the captured scene and many input pixels to the camera are modified. 
The original and attacked scenes are shown in Figure \ref{2D attack}.

Through this attack, the probability of \texttt{horse} was decreased from 98\% to 22\%.

\subsection{3D Presentation} 
\begin{figure*}[]
	\centering
	\begin{subfigure}[]{0.45\textwidth}
		\centering
		\includegraphics[width=.6\textwidth]{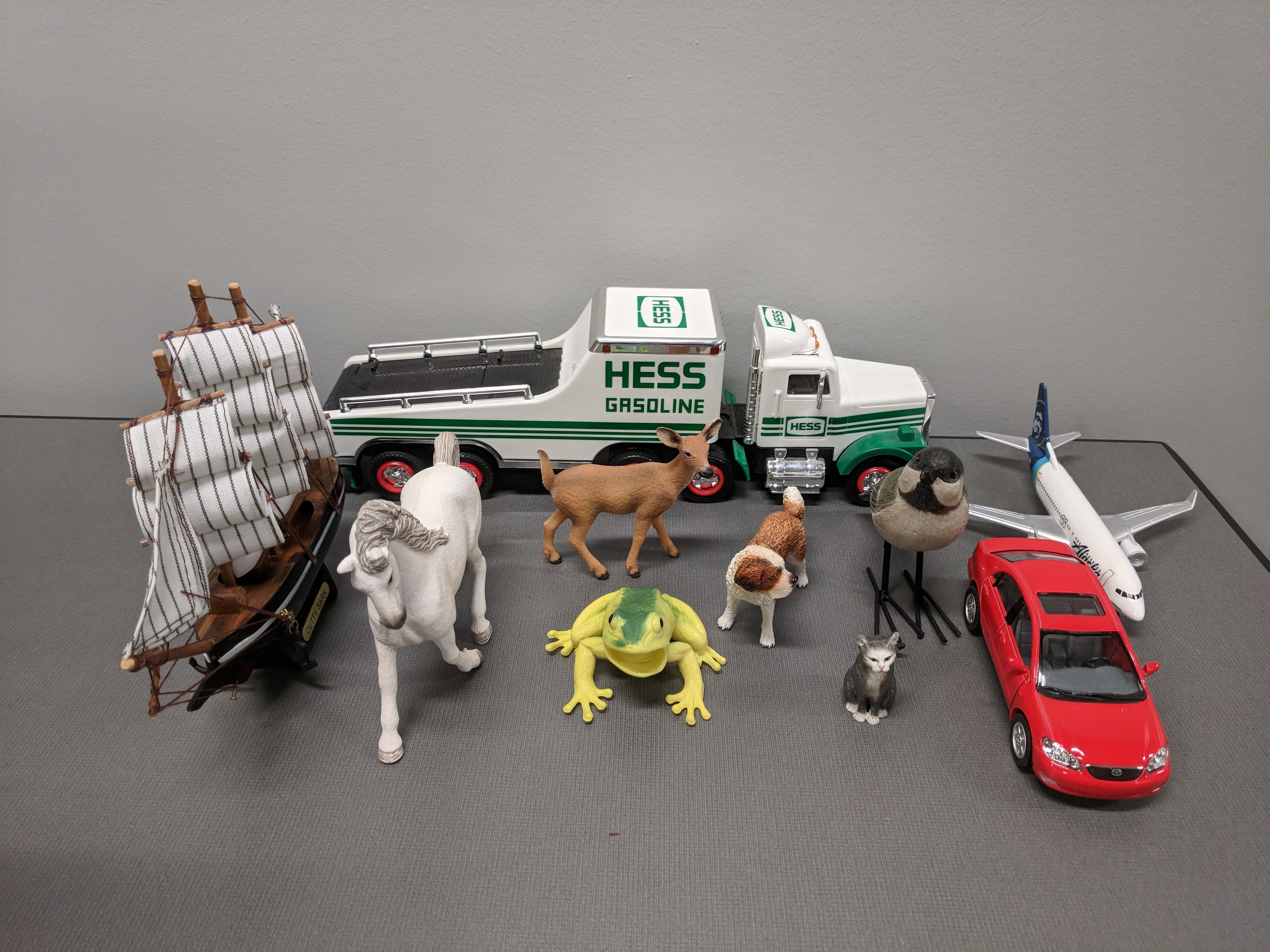}
		\caption{The toy figurines used to represent the CIFAR classes.}\label{fig:cifarSet}
	\end{subfigure}
	~
	\begin{subfigure}[]{0.45\textwidth}
		\centering
		\includegraphics[width=.45\textwidth]{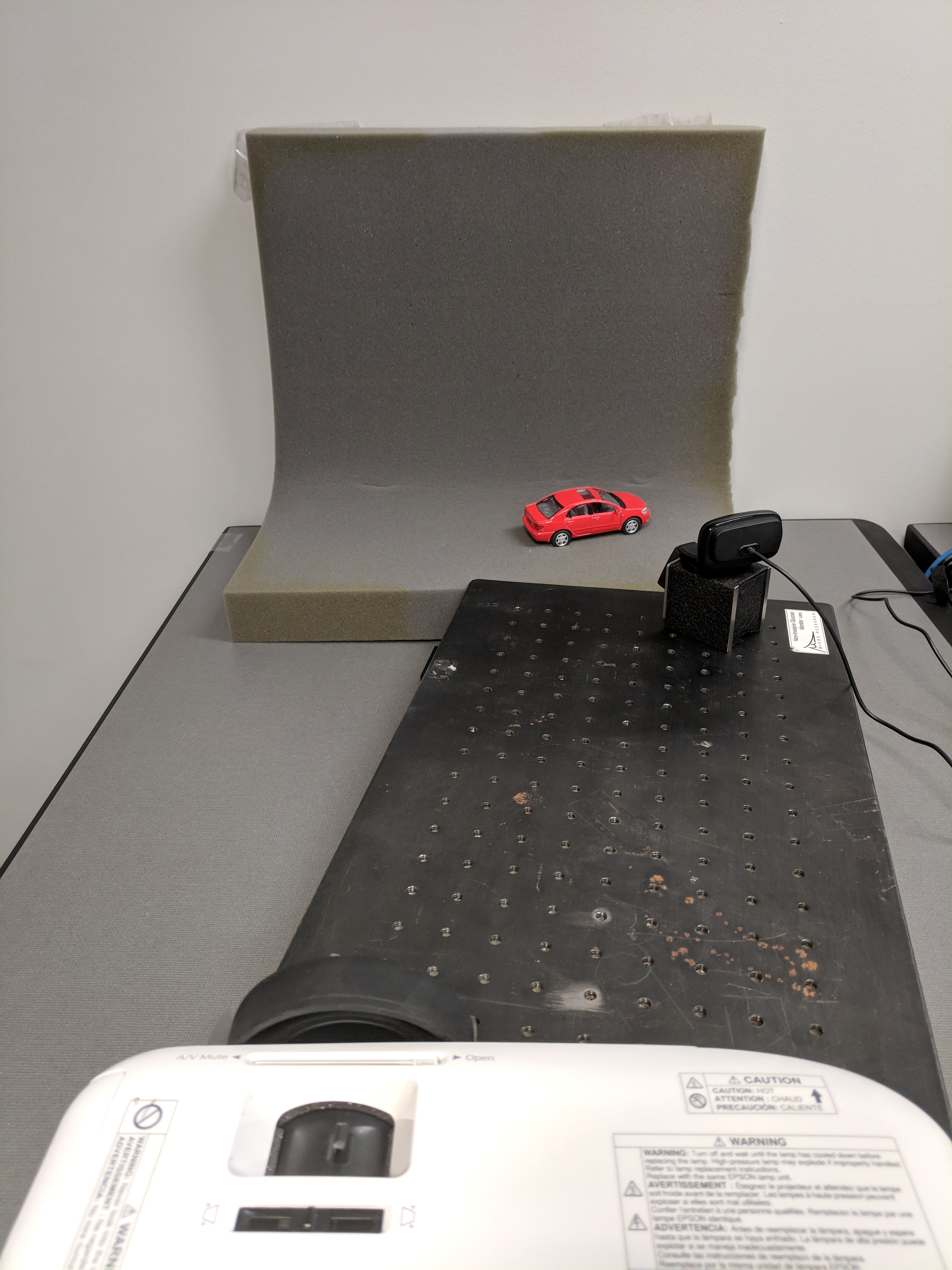}
		\caption{Physical setup demonstrating relative position of projector, camera, object, and lighting control.}\label{fig:tabletop}
	\end{subfigure}
	\caption{Experimental setup and figurines for second phase experiments with 3D presentation.}\label{3Dsetup2}
\end{figure*}

To demonstrate the potential for light based attacks, we extended the 2D methodology to a 3D scene in two experimental phases. 
First, we placed a toy car in the field of view of the web camera to capture the scene.
To perform the attack, the projector iteratively applies the same adversarial noise procedure to the 3D physical scene and the same ResNet38 model is used for evaluation. 
The object probabilities for the original scene were 89\% \texttt{automobile} and 11\% \texttt{truck}. The attacked scene probabilities were 43 \% \texttt{automobile} and 57\% \texttt{truck}.   
The second phase of experiments was designed to improve the repeatability and confidence of the initial demonstration.
Results are expanded to evaluate all 10 CIFAR classes: \texttt{airplane}, \texttt{automobile}, \texttt{bird}, \texttt{cat}, \texttt{deer}, \texttt{dog}, \texttt{frog}, \texttt{horse}, \texttt{ship}, \texttt{truck}.  
The figurines used for each of these classes are shown in Figure \ref{fig:cifarSet}. The yellow car in phase 1 was not available and was replaced with a red car in phase 2. 

Rotation invariance is important for interpreting the presented experimental setup. 
This impacts our data collection because we observed in a baseline condition, with no added light, the distance to the camera and object orientation yielded highly variable classification results.
We tested four experimental conditions: ambient light, white light from the projector, white light with a randomly located pixel in the 32x32 grid, and differential evolution process to control color and location of one pixel in a 32x32 white grid. We observed classification variability in the physical scene when no modifications were applied. 
For this reason we introduced some lighting controls which observationally provided a significantly more stable baseline classification. 
Three physical modifications were made. 
The projected background color was changed from black to white to provide more uniformity to the scene. 
We used a foam block to minimize stray reflections caused by the projector.
Additionally we used a neutral density filter to scale the light intensity.  
To verify stability, we collected twenty image captures of each test condition, and 200 for differential evolution (50 population sample and 4 evolution phases). 
 
Reproducibility of the physical placement of each object in the scene is imprecise, thus each test condition was collected in sequence without any disturbance (besides light). 
An unrecorded calibration phase was used to reposition the object for a maximum baseline classification score before the recorded baseline and light projected data was collected. 
For each class and test condition, we report the mean, median, standard deviation, variance, minimum, maximum, $\Delta mean$ and $\Delta median$. 
The $\Delta mean$  and $\Delta median$ are the computation of the reduction in probability score for the given attack type relative to baseline. 
Larger $\Delta$ numbers represent more powerful decrease in the true class probability. 
All scores are reported in Table \ref{massiveTable}.

Interpreting the table yields one immediate observation: some examples (\texttt{Automobile}, \texttt{Bird}, \texttt{Horse}, \texttt{Ship}) are invariant to the light attack, consistently being identified as the true class at 100\% (within rounding error) while other classes (\texttt{Airplane}, \texttt{Cat}, \texttt{Deer}, \texttt{Dog}, \texttt{Frog}, and \texttt{Truck}) have varying degrees of susceptibility.
It is unclear whether these differences are inherent in the classes themselves, or to the particular figurines we chose.
As one might expect with a research classifier, there is a high degree of variability based on the particular example.
We incremented the complexity of light attack from pure white light, random square, and differential evolution, to assess if there was something unique in the more sophisticated attack, or if it was merely the addition of light, or a pattern, that was causing the observed decrease in classification.  
In many cases, the simple addition of white light is as effective as the other attacks.  
For example, the mean airplane class was decreased from 1.000 to 0.151, with only the addition of white light. 
The corresponding trials with random and differential evolution light patterns yielded only slightly stronger attacks, with 0.113 and 0.133 mean scores respectively. However, the decline is noteworthy, independent of sophistication.

\begin{figure*}[htp]
	\centering
	\begin{subfigure}[]{0.45\textwidth}
		\centering
		\includegraphics[width=0.6\textwidth]{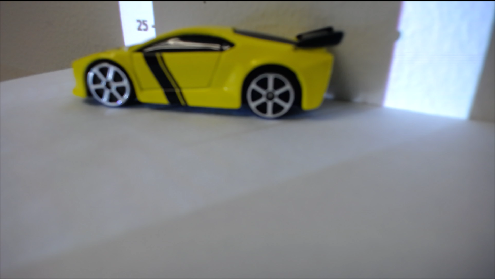}
		\caption{The 3D scene without any adversarial attack. }\label{fig:car}
	\end{subfigure}
	~
	\begin{subfigure}[]{0.45\textwidth}
		\centering
		\includegraphics[width=0.6\textwidth]{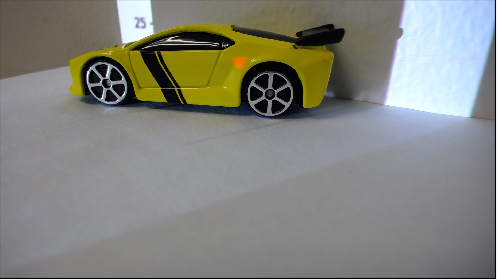}
		\caption{The 3D scene with adversarial attack. }\label{fig:car-attack}
	\end{subfigure}
	\caption{Images demonstrating light based attack on 3D physical presentation}\label{ 3Dcapture}
\end{figure*}

\begin{figure*}[htp]
	\centering
	\begin{subfigure}[]{0.45\textwidth}
		\centering
		\includegraphics[width=0.6\textwidth]{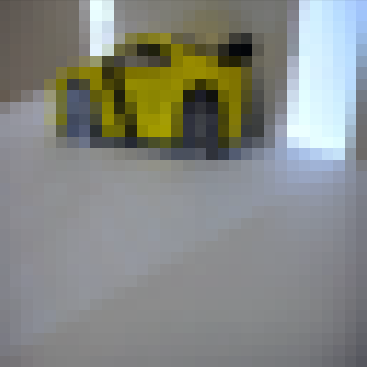}
		\caption{Downsampled image without any adversarial attack. }\label{fig:car32}
	\end{subfigure}
	~
	\begin{subfigure}[]{0.45\textwidth}
		\centering
		\includegraphics[width=0.6\textwidth]{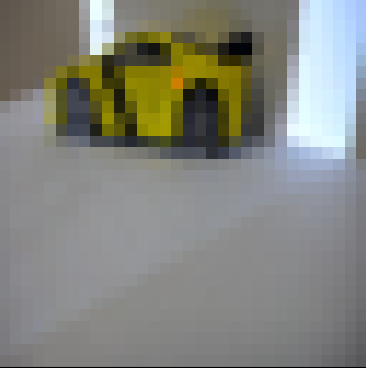}
		\caption{Downsampled image with adversarial attack. }\label{fig:car-attack32}
	\end{subfigure}
	\caption{Downsampled images demonstrating light based attack on 3D physical representation}\label{3DDsample}
\end{figure*}

\begin{table*}[t]
	\begin{tabular*}{\columnwidth}{|c|c|c|c|c|c|c|c|c|c|}
		\cline{1-10}
		CIFAR Class & Experiment Condition & Mean & Median & SD & Var & Min & Max & $\Delta$ Mean & $\Delta$ Median \\ 
		\cline{1-10}
		Airplane & Baseline & 1.000 & 1.000 & .000 & .000 & 1.000 & 1.000 & .000 & .000 \\ 
		\cline{2-10}
		& White Light & .151 & .101 & .198 & .039 & .017 & .997 & .849 & .899 \\ 
		\cline{2-10}
		& Random & .114 & .105 & .088 & .008 & .022 & .445 & .886 & .895 \\ 
		\cline{2-10}
		& Diff Evolution & .133 & .112 & .087 & .007 & .014 & .459 & .867 & .888 \\ 
		\cline{1-10}
		Automobile & Baseline & 1.000 & 1.000 & .000 & .000 & 1.000 & 1.000 & .000 & .000 \\ 
		\cline{2-10}
		& White Light & 1.000 & 1.000 & .000 & .000 & .999 & 1.000 & .000 & .000 \\ 
		\cline{2-10}
		& Random & 1.000 & 1.000 & .000 & .000 & .999 & 1.000 & .000 & .000 \\ 
		\cline{2-10}
		& Diff Evolution & 1.000 & 1.000 & .000 & .000 & 1.000 & 1.000 & .000 & .000 \\ 
		\cline{1-10}
		Bird & Baseline & 1.000 & 1.000 & .000 & .000 & 1.000 & 1.000 & .000 & .000 \\ 
		\cline{2-10}
		& White Light & 1.000 & 1.000 & .002 & .000 & .993 & 1.000 & .000 & .000 \\ 
		\cline{2-10}
		& Random & 1.000 & 1.000 & .000 & .000 & 1.000 & 1.000 & .000 & .000 \\ 
		\cline{2-10}
		& Diff Evolution & 1.000 & 1.000 & .000 & .000 & .999 & 1.000 & .000 & .000 \\ 
		\cline{1-10}
		Cat & Baseline & .990 & .991 & .004 & .000 & .979 & .996 & .000 & .000 \\ 
		\cline{2-10}
		& White Light & .009 & .008 & .005 & .000 & .000 & .020 & .981 & .983 \\ 
		\cline{2-10}
		& Random & .011 & .007 & .012 & .000 & .001 & .047 & .979 & .984 \\ 
		\cline{2-10}
		& Diff Evolution & .023 & .017 & .019 & .000 & .002 & .124 & .967 & .974 \\ 
		\cline{1-10}
		Deer & Baseline & .999 & .999 & .000 & .000 & .999 & 1.000 & .000 & .000 \\ 
		\cline{2-10}
		& White Light & .516 & .516 & .145 & .021 & .242 & .997 & .483 & .483 \\ 
		\cline{2-10}
		& Random  & .545 & .507 & .155 & .024 & .327 & .871 & .454 & .492 \\ 
		\cline{2-10}
		& Diff Evolution & .473 & .467 & .130 & .017 & .144 & .829 & .526 & .532 \\ 
		\cline{1-10}
		Dog & Baseline & .993 & .993 & .003 & .000 & .986 & .996 & .000 & .000 \\ 
		\cline{2-10}
		& White Light & .512 & .499 & .088 & .008 & .390 & .695 & .481 & .494 \\ 
		\cline{2-10}
		& Random & .482 & .497 & .123 & .015 & .136 & .753 & .511 & .496 \\ 
		\cline{2-10}
		& Diff Evolution & .386 & .388 & .088 & .008 & .123 & .601 & .606 & .605 \\ 
		\cline{1-10}
		Frog & Baseline & .888 & .888 & .025 & .001 & .842 & .933 & .000 & .000 \\ 
		\cline{2-10}
		& White Light & .008 & .008 & .003 & .000 & .000 & .015 & .881 & .880 \\ 
		\cline{2-10}
		& Random & .030 & .011 & .076 & .006 & .004 & .360 & .858 & .877 \\ 
		\cline{2-10}
		& Diff Evolution & .071 & .038 & .093 & .009 & .005 & .576 & .817 & .849 \\ 
		\cline{1-10}
		Horse & Baseline & 1.000 & 1.000 & .000 & .000 & 1.000 & 1.000 & .000 & .000 \\ 
		\cline{2-10}
		& White Light & .999 & 1.000 & .001 & .000 & .993 & 1.000 & .000 & .000 \\ 
		\cline{2-10}
		& Random & 1.000 & 1.000 & .000 & .000 & 1.000 & 1.000 & .000 & .000 \\ 
		\cline{2-10}
		& Diff Evolution & 1.000 & 1.000 & .000 & .000 & 1.000 & 1.000 & .000 & .000 \\ 
		\cline{1-10}
		Ship & Baseline & 1.000 & 1.000 & .000 & .000 & 1.000 & 1.000 & .000 & .000 \\ 
		\cline{2-10}
		& White Light & 1.000 & 1.000 & .000 & .000 & 1.000 & 1.000 & .000 & .000 \\ 
		\cline{2-10}
		& Random & 1.000 & 1.000 & .000 & .000 & 1.000 & 1.000 & .000 & .000 \\ 
		\cline{2-10}
		& Diff Evolution & 1.000 & 1.000 & .000 & .000 & 1.000 & 1.000 & .000 & .000 \\ 
		\cline{1-10}
		Truck & Baseline & 1.000 & 1.000 & .000 & .000 & 1.000 & 1.000 & .000 & .000 \\ 
		\cline{2-10}
		& White Light & .832 & .832 & .052 & .003 & .729 & 1.000 & .168 & .168 \\ 
		\cline{2-10}
		& Random & .818 & .819 & .072 & .005 & .634 & .970 & .182 & .180 \\ 
		\cline{2-10}
		& Diff Evolution & .826 & .839 & .088 & .008 & .507 & .949 & .174 & .161 \\ 
		\cline{1-10}
	\end{tabular*} 
\caption{Classification statistics for baseline and attacked CIFAR figures.}\label{massiveTable}
\end{table*}

%
%
%
%
%

\section{Discussion}
Physical attacks on machine learning systems could be applied in a wide range of security domains. 
The literature has primarily discussed the safety of road signs and autonomous driving \cite{Eykholt2017,chen2018robust}, however other security applications may also be impacted. 
An adversary may be trying to hide themselves or physical ties to illegal activities to evade law enforcement (e.g. knives/weapons, contraband, narcotics manufacturing, etc). 
Any AI to be deployed for law-enforcement applications needs to be robust in an adversarial environment where physical obfuscation could be employed.  
Light based attacks:
\begin{itemize}
	\item Can perform targeted and non-targeted attacks.
	\item Do not modify physical object in a permanent way.
	\item Can be a transient effect occurring at specified times. 
\end{itemize}
This work aims to be a first step towards understanding the abilities and limitations of such physical attacks. 
We picked a relatively easy first target to verify the possibility and plan to extend this to more complex physical scenarios and classification models. 

We chose to attack the CIFAR-10 framework in a manner similar to what was demonstrated in the original single pixel attack \cite{su2017one}.
This framework is an easier target because it is a low resolution, low parameter model. To assess the robustness of stronger models, a ResNet50 classifier trained on ImageNet was also used to evaluate all of the collected images. Because of a lack of corresponding true class identification, scores are not reported, but it was observed that the top1 class prediction was shifted with the addition of light based attacks. 

There is also a closed world assumption of 10 relatively dissimilar classes, where the probability of all classes sums to one. 
When a misclassification occurs, it tends to be more outlandish than it could otherwise be. 
For example, \texttt{rose} and \texttt{tulip} might be a more forgiving mistake than \texttt{frog} and \texttt{airplane} but in the CIFAR closed world framework, the model is limited to the 10 known classes. 

In our attack on the 3D presentation, the true class was correctly identified as \texttt{car} when no attack was present.
By applying the adversarial light attack, we were able to decrease the confidence of \texttt{car} from 89\% to 43\%, and instead predict \texttt{truck} with 57\% probability. 
We would not identify this as a 3D attack because we had a fixed orientation  between the camera, projector, and object. 
In this example, the single square attack is visually perceptible but transient. 
However, the notion of human perception is not as simple as an $L_\infty$ distance in pixel space. 
This is highlighted by the fact that consecutive video frames can be significantly mis-classified by top performing image classification systems \cite{Zheng2016}.  
Images that are imperceptibly different can have large distance in pixel or feature space, and images that are perceptually different can be close. 

A key topic that needs further understanding is why the extreme variability in class identification.  
One potential explanation is the degree of self similarity within a class, and training data bias. 
For example, the horse images in the training data, are potentially all self similar and also closely match the example figurine.  The variation between different types of horses is likely smaller than the visual difference between different breeds of dogs. 

Another possible explanation is the scale or percentage of the scene that the object occupies.  
Most of the classes which were susceptible to attack were relatively small.  
The notable exception was the truck which was actually the largest figure used for data, yet was still susceptible to misclassification errors with the addition of light. 

There are a several important constraints present when crafting a light based physical attack that are unconstrained in a digital attack. 
Specifically, light is always an additive noise and turning a dark color to white with the addition of light is impossible. 
The angle of projection and the texture of the scene may impact the colors reflected to the camera. 
The camera itself will introduce color balance changes as it adjusts to the adversarial addition of light. 
Even a fully manual camera will always have CCD shot noise, which is a function of shutter speed and temperature, that could influence the success or failure of a light based attack. 
The projected pixel was not constrained to overlap the target object, and would appear in the background. 
Empirically, these single pixel projections onto the background of an image could significantly change classifier predictions.




\section{Conclusion and Future Work}
The presented work is an empirical demonstration of light based attacks on deep learning based object recognition systems.  
Adversarial machine learning research has emphasized attacks against deep learning architectures, however it has been observed that other models are equally susceptible to attack and that adversarial examples often transfer between model types \cite{Papernot2016}. 
The empirical demonstration of light based attack was against a deep learning architecture. 
However, based on this prior work, it is likely that it could be demonstrated against other model types.

We plan on conducting experiments with higher resolution and more robust classifiers and more subtle manipulations. 
We believe that more targeted optimization approaches that initially focus on sensitive image areas will likely lead to faster identification of successful attacks.
We expect light based attacks could use more complex projected textures and take advantage of 3D geometry. 
Presented results clearly show light has the potential to be another avenue of adversarial attack in the physical domain. 

\section{Acknowledgments}
The research described in this paper is part of the Analysis in Motion Initiative at Pacific Northwest National Laboratory; conducted under the Laboratory Directed Research and Development Program at PNNL, a multi-program national laboratory operated by Battelle for the U.S. Department of Energy. The authors are especially grateful to Mark Greaves, Artem Yankov, Sean Zabriskie, Michael Henry, Jeremiah Rounds, Court Corley, Nathan Hodas, Will Koella and our Quickstarter supporters.  
\\
\\

\bibliographystyle{alec18}
\bibliography{ALEC-2018}

\begin{thebibliography}{}

\bibitem[\protect\citeauthoryear{Athalye and
  Sutskever}{2017}]{athalye2017synthesizing}
Athalye, A., and Sutskever, I.
\newblock 2017.
\newblock Synthesizing robust adversarial examples.
\newblock {\em arXiv preprint arXiv:1707.07397}.

\bibitem[\protect\citeauthoryear{Athalye, Carlini, and
  Wagner}{2018}]{athalye2018obfuscated}
Athalye, A.; Carlini, N.; and Wagner, D.
\newblock 2018.
\newblock Obfuscated gradients give a false sense of security: Circumventing
  defenses to adversarial examples.
\newblock {\em arXiv preprint arXiv:1802.00420}.

\bibitem[\protect\citeauthoryear{Brown \bgroup et al\mbox.\egroup
  }{2017}]{brown2017adversarial}
Brown, T.~B.; Man{\'e}, D.; Roy, A.; Abadi, M.; and Gilmer, J.
\newblock 2017.
\newblock Adversarial patch.
\newblock {\em arXiv preprint arXiv:1712.09665}.

\bibitem[\protect\citeauthoryear{Chen \bgroup et al\mbox.\egroup
  }{2018}]{chen2018robust}
Chen, S.-T.; Cornelius, C.; Martin, J.; and Chau, D.~H.
\newblock 2018.
\newblock Robust physical adversarial attack on faster r-cnn object detector.
\newblock {\em arXiv preprint arXiv:1804.05810}.

\bibitem[\protect\citeauthoryear{Cohen and
  Welling}{2014}]{cohen2014transformation}
Cohen, T.~S., and Welling, M.
\newblock 2014.
\newblock Transformation properties of learned visual representations.
\newblock {\em arXiv preprint arXiv:1412.7659}.

\bibitem[\protect\citeauthoryear{Eykholt \bgroup et al\mbox.\egroup
  }{2017}]{Eykholt2017}
Eykholt, K.; Evtimov, I.; Fernandes, E.; Li, B.; Rahmati, A.; Xiao, C.;
  Prakash, A.; Kohno, T.; and Song, D.
\newblock 2017.
\newblock {Robust Physical-World Attacks on Deep Learning Models}.

\bibitem[\protect\citeauthoryear{Feinman \bgroup et al\mbox.\egroup
  }{2017}]{feinman2017detecting}
Feinman, R.; Curtin, R.~R.; Shintre, S.; and Gardner, A.~B.
\newblock 2017.
\newblock Detecting adversarial samples from artifacts.
\newblock {\em arXiv preprint arXiv:1703.00410}.

\bibitem[\protect\citeauthoryear{Jaderberg \bgroup et al\mbox.\egroup
  }{2015}]{jaderberg2015spatial}
Jaderberg, M.; Simonyan, K.; Zisserman, A.; et~al.
\newblock 2015.
\newblock Spatial transformer networks.
\newblock In {\em Advances in neural information processing systems},
  2017--2025.

\bibitem[\protect\citeauthoryear{Kurakin, Goodfellow, and
  Bengio}{2016}]{Kurakin2016}
Kurakin, A.; Goodfellow, I.; and Bengio, S.
\newblock 2016.
\newblock {Adversarial examples in the physical world}.
\newblock {\em Arxiv} (c):1--15.

\bibitem[\protect\citeauthoryear{LeCun, Huang, and
  Bottou}{2004}]{lecun2004learning}
LeCun, Y.; Huang, F.~J.; and Bottou, L.
\newblock 2004.
\newblock Learning methods for generic object recognition with invariance to
  pose and lighting.
\newblock In {\em Computer Vision and Pattern Recognition, 2004. CVPR 2004.
  Proceedings of the 2004 IEEE Computer Society Conference on}, volume~2,
  II--104.
\newblock IEEE.

\bibitem[\protect\citeauthoryear{Papernot, McDaniel, and
  Goodfellow}{2016}]{Papernot2016}
Papernot, N.; McDaniel, P.; and Goodfellow, I.
\newblock 2016.
\newblock {Transferability in Machine Learning: from Phenomena to Black-Box
  Attacks using Adversarial Samples}.

\bibitem[\protect\citeauthoryear{Qi \bgroup et al\mbox.\egroup
  }{2017}]{qi2017pointnet}
Qi, C.~R.; Su, H.; Mo, K.; and Guibas, L.~J.
\newblock 2017.
\newblock Pointnet: Deep learning on point sets for 3d classification and
  segmentation.
\newblock {\em Proc. Computer Vision and Pattern Recognition (CVPR), IEEE}
  1(2):4.

\bibitem[\protect\citeauthoryear{Sharif \bgroup et al\mbox.\egroup
  }{2016}]{sharif2016accessorize}
Sharif, M.; Bhagavatula, S.; Bauer, L.; and Reiter, M.~K.
\newblock 2016.
\newblock Accessorize to a crime: Real and stealthy attacks on state-of-the-art
  face recognition.
\newblock In {\em Proceedings of the 2016 ACM SIGSAC Conference on Computer and
  Communications Security},  1528--1540.
\newblock ACM.

\bibitem[\protect\citeauthoryear{Su, Vargas, and Kouichi}{2017}]{su2017one}
Su, J.; Vargas, D.~V.; and Kouichi, S.
\newblock 2017.
\newblock One pixel attack for fooling deep neural networks.
\newblock {\em arXiv preprint arXiv:1710.08864}.

\bibitem[\protect\citeauthoryear{Szegedy \bgroup et al\mbox.\egroup
  }{2013}]{szegedy2013intriguing}
Szegedy, C.; Zaremba, W.; Sutskever, I.; Bruna, J.; Erhan, D.; Goodfellow, I.;
  and Fergus, R.
\newblock 2013.
\newblock Intriguing properties of neural networks.
\newblock {\em arXiv preprint arXiv:1312.6199}.

\bibitem[\protect\citeauthoryear{Yamada, Gohshi, and
  Echizen}{2013}]{yamada2013privacy}
Yamada, T.; Gohshi, S.; and Echizen, I.
\newblock 2013.
\newblock Privacy visor: Method for preventing face image detection by using
  differences in human and device sensitivity.
\newblock In {\em IFIP International Conference on Communications and
  Multimedia Security},  152--161.
\newblock Springer.

\bibitem[\protect\citeauthoryear{Zheng \bgroup et al\mbox.\egroup
  }{2016}]{Zheng2016}
Zheng, S.; Song, Y.; Leung, T.; and Goodfellow, I.
\newblock 2016.
\newblock {Improving the Robustness of Deep Neural Networks via Stability
  Training}.

\end{thebibliography}

\end{document}